\documentclass[10pt,twocolumn,letterpaper]{article}

\usepackage[pagenumbers]{cvpr} 
\usepackage{graphicx}
\usepackage{amsmath}
\usepackage{amssymb}
\usepackage{booktabs}
\usepackage{cite}
\usepackage[ruled,vlined]{algorithm2e}
\usepackage{pifont}
\usepackage{float}
\usepackage[pagebackref,breaklinks,colorlinks]{hyperref}
\usepackage{multirow}
\usepackage[table]{xcolor}
\definecolor{lightpurple}{RGB}{200, 230, 245}
\definecolor{lightblue}{RGB}{230, 245, 255}
\usepackage[capitalize]{cleveref}
\crefname{section}{Sec.}{Secs.}
\Crefname{section}{Section}{Sections}
\Crefname{table}{Table}{Tables}
\crefname{table}{Tab.}{Tabs.}

\def\cvprPaperID{238} 
\def\confName{CVPR}
\def\confYear{2025}

\begin{document}

\title{Beyond Single Prompts: Synergistic Fusion and Arrangement for VICL}

\author{
    Wenwen Liao\textsuperscript{1} \quad
    Jianbo Yu\textsuperscript{2}\thanks{Corresponding author: {\tt\small jb\_yu@fudan.edu.cn}} \quad
    Yuansong Wang\textsuperscript{3} \quad
    Shifu Yan\textsuperscript{4} \quad
    Xiaofeng Yang\textsuperscript{2} \\[2mm]
    \textsuperscript{1}College of Intelligent Robotics and Advance Manufacturing, Fudan University\\
    \textsuperscript{2}School of Microelectronics, Fudan University\\
    \textsuperscript{3}Tsinghua Shenzhen International Graduate School, Tsinghua University\\
    \textsuperscript{4}ByteDance Ltd.\\
    {\tt\small wwliao24@m.fudan.edu.cn, jb\_yu@fudan.edu.cn} 
}

\twocolumn[{
\maketitle
\begin{center}
    \captionsetup{type=figure}
    \includegraphics[width=1\textwidth]{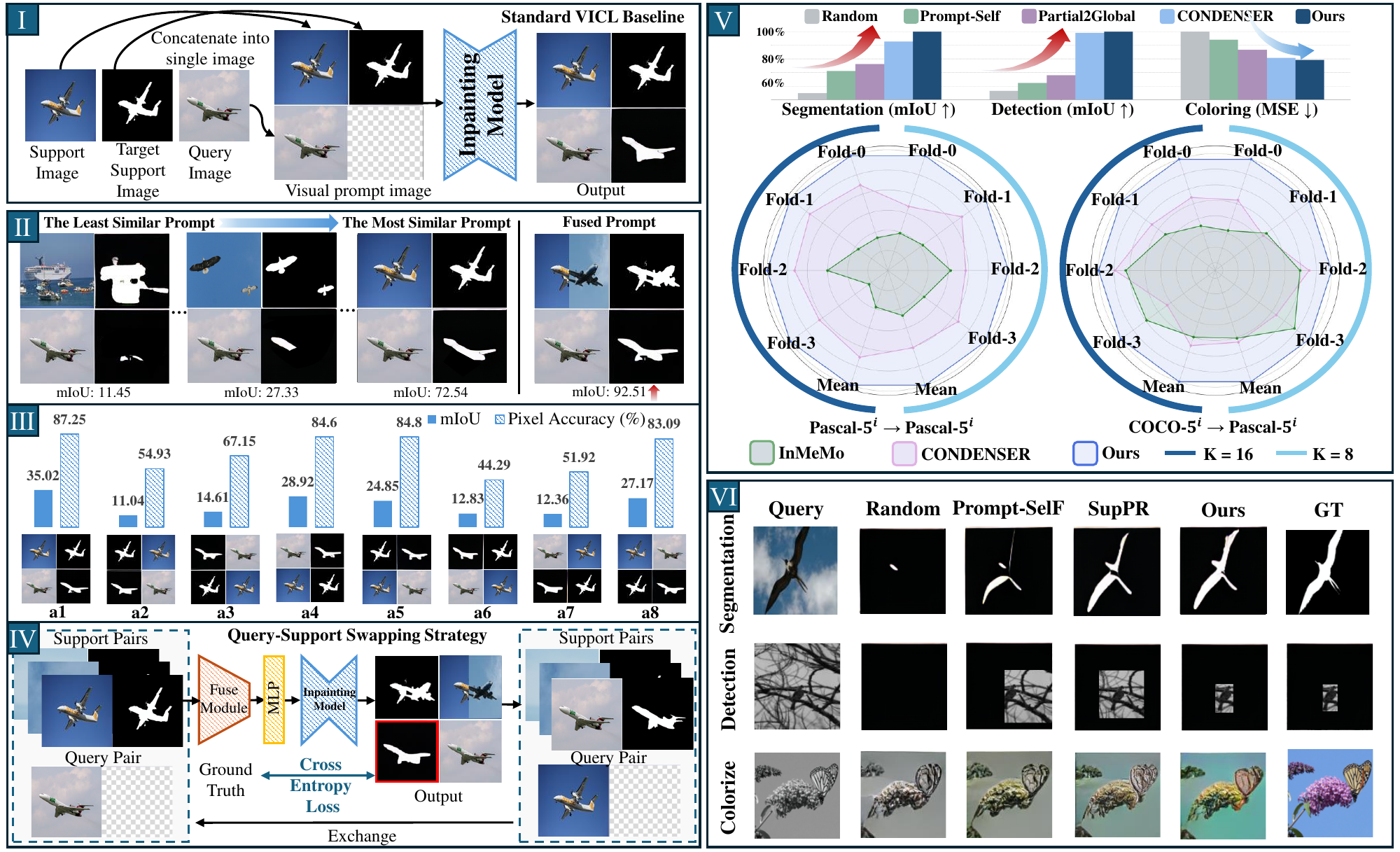}
    \captionof{figure}{\textbf{(I)} The standard VICL pipeline, which essentially performs image inpainting guided by support prompts. \textbf{(II)} One critical factor of VICL, \emph{prompt selection}. While more similar prompts yield better results, our fused prompt surpasses even the most similar one.
    \textbf{(III)} Another critical factor, \emph{prompt arrangement}. Varying layouts influence the model inpainting in distinct ways. \textbf{(IV)} Our proposed joint fine-tuning strategy. \textbf{(V, VI)} Quantitative and qualitative results demonstrating our model's state-of-the-art performance and generalization.}

    \label{top}
\end{center}
}]

\begin{abstract}
Vision In-Context Learning (VICL) enables inpainting models to quickly adapt to new visual tasks from only a few prompts. However, existing methods suffer from two key issues: (1) selecting only the most similar prompt discards complementary cues from other high-quality prompts; and (2) failing to exploit the structured information implied by different prompt arrangements.

We propose an end-to-end VICL framework to overcome these limitations. Firstly, an \textbf{adaptive Fusion Module} aggregates critical patterns and annotations from multiple prompts to form more precise contextual prompts. Secondly, we introduce \textbf{arrangement-specific lightweight MLPs} to decouple layout priors from the core model, while minimally affecting the overall model. In addition, an \textbf{bidirectional fine-tuning mechanism} swaps the roles of query and prompt, encouraging the model to reconstruct the original prompt from fused context and thus enhancing collaboration between the fusion module and the inpainting model. Experiments on foreground segmentation, single-object detection, and image colorization demonstrate superior results and strong cross-task generalization of our method. Code will be released upon acceptance.

\end{abstract}

\section{Introduction}
\label{sec:intro}

In-Context Learning (ICL), as an emerging learning paradigm, is first revealed in the GPT-3~\cite{brown2020language}. 
It endows models with the remarkable ability to rapidly adapt to and perform novel downstream tasks 
by leveraging only a few prompts provided within user prompts. 
Following its great success in the field of Natural Language Processing (NLP)~\cite{hao2022language,brown2020language,alayrac2022flamingo}, 
this concept has been extended to computer vision, giving rise to 
\emph{Visual In-Context Learning} (VICL), which aims to build general-purpose models for vision tasks~\cite{bar2022visual,wang2023seggpt,bai2024sequential,bahng2022exploring}.

At present, mainstream VICL approaches elegantly unify diverse vision tasks 
into a formulation of image inpainting or completion. 
As illustrated in Figure~\ref{top}(I), VICL concatenates ``input--output'' support prompts, which describe a specific task, with a query image into a composite image grid. An Inpainting Model then predicts and ``fills in'' the blank regions of the grid such that the completed content 
remains consistent with the provided task demonstrations. 
With the guidance of only a few prompts, a pretrained vision model can be flexibly adapted to a wide range of downstream applications, such as image segmentation, object detection, and image colorization.

However, as noted in~\cite{sun2025exploring}, the final performance of the model is highly sensitive to two interrelated dimensions: 
\emph{prompt selection} and \emph{prompt arrangement}. 
The quality of prompts directly determines the depth of task understanding by the model 
(as illustrated in Figure~\ref{top}(II)), making high-quality demonstrations essential and driving extensive research on prompt selection. Partial2Global~\cite{xu2024towards} ranks prompts via list-wise comparison and consistency-aware aggregation, while~\cite{zhang2023makes} provides two weight-free methods: either through nearest-neighbor retrieval or a supervised selector trained to maximize in-context performance. However, these ranking-based winner-takes-all strategies have inherent limitations, as it selects only the single best prompt while discarding potentially valuable complementary information contained in other high-quality prompts. Although CONDENSER \cite{wang2025embracing} pioneers to explore the benefits of fusing multiple prompts, their modular design separates fusion from inpainting, leaving the generator as a passive receiver, rather than an active participant in the fusion process.

In addition to prompt selection, the geometric arrangement in the image grids is equally critical (as shown in Figure~\ref{top}(III)), as different arrangements guide knowledge use and lead to performance variations. However, most existing methods overlook this factor and still follow sequential fusion pipelines (pattern “a1”), thus failing to exploit the rich potential encoded in various arrangements. Prompt-SelF~\cite{sun2025exploring} generates multiple predictions by concatenating the query image and the prompt image in eight different ways and feeding them into the same frozen vision model, after which the results are ensembled by voting to enhance the final prediction accuracy. Although this strategy relies on a single model that is fundamentally “blind" to prompt arrangements it is fed, inherently limiting its potential.

To address these shortcomings, this paper proposes a novel, end-to-end framework. In contrast to existing methods, our framework jointly optimizes the fusion and generation processes. It also incorporates arrangement-aware Multilayer Perceptrons (MLPs), resulting in a more powerful representation than can be achieved with sequential fusion approaches. Specifically, to leverage multiple high-quality prompts, we introduce a Fusion Module for adaptive prompt fusion. This module is engineered to identify relevant image patterns and their corresponding annotations within the candidates, constructing a tailored reference for the local regions of each query image. This mechanism facilitates the extraction of the most pertinent fine-grained context for the final fusion prompt.
Second, to learn arrangement-specific knowledge with minimal impact on the core model, we introduce a set of MLPs for prompt arrangements. Each MLP is independently trained to achieve the best performance under its specific layout, after which the results across layouts are compared to select a few arrangements with better performance for the next step.
Additionally, to enhance the general capability of the entire model, we implement a fine-tuning strategy (as shown in Figure~\ref{top}(IV)) where one of the support prompt pairs is swapped with the query and the generated image. This allows us to jointly train the whole model, thereby further enhancing their synergy. The best performing arrangement is then selected as the final model.

Extensive experiments on multiple benchmark tasks demonstrate the superior performance of our model.
As shown in Figure~\ref{top}(V), our method significantly outperforms existing models, across segmentation, detection, and coloring tasks.
This advantage is pronounced not only in the in-domain setting (Pascal-5$^i$ $\rightarrow$ Pascal-5$^i$) but also in the more challenging cross-domain scenario (COCO-5$^i$ $\rightarrow$ Pascal-5$^i$), where our model maintains a comprehensive and substantial lead.
Furthermore, the qualitative results in Figure~\ref{top}(VI) intuitively showcase the visual superiority of our approach. This further highlights our model's powerful generalization capabilities and its precise contextual understanding and fusion abilities.

To sum up, this paper's contributions are:

\begin{itemize}
    \item \textbf{Novel End-to-End VICL Framework:} We propose an end-to-end framework that simultaneously addresses the VICL limitations of context fusion and prompt arrangement.

    \item \textbf{Adaptive Prompt Fusion Module:} We propose a Fusion Module that adaptively fuses fine-grained context from multiple prompts to create rich, localized references for the query.
    
    \item \textbf{Geometry-Semantic Decoupling Module:} A set of lightweight MLPs are assigned respectively to each prompt layout to embed its geometric structure, explicitly separating spatial encoding from semantic processing, avoiding interference with the core model.
    
    \item \textbf{Synergistic Training Strategy:} A query-support swapping strategy enables joint optimization of the Fusion and Inpainting modules, enhancing both contextual fusion and reconstruction performance.

    \item \textbf{Comprehensive Experimental Validations:} Extensive benchmarks demonstrate strong performance and generalization.
\end{itemize}


\begin{figure}
  \centering
  \includegraphics[width=\linewidth]{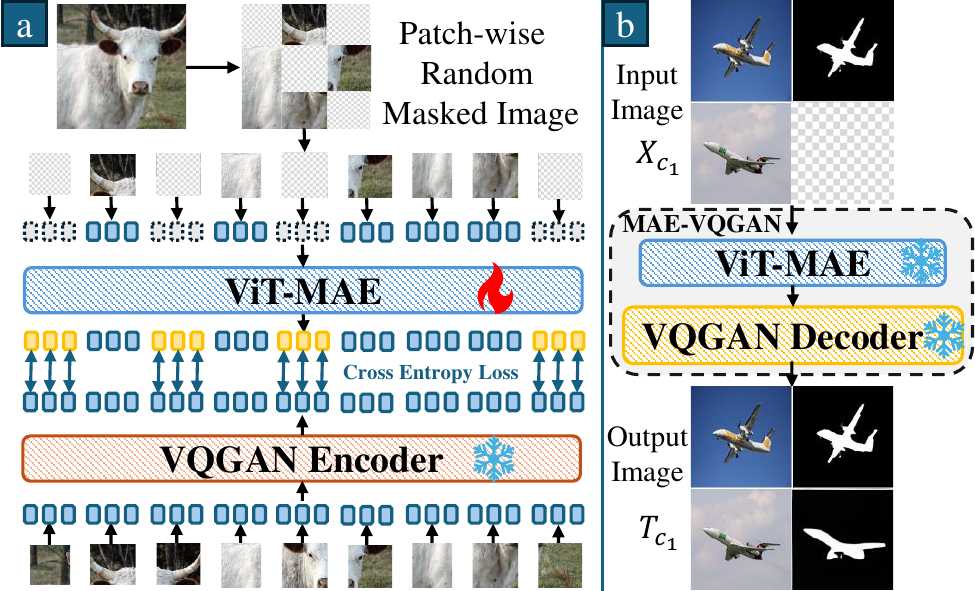} 
  \caption{\textbf{MAE-VQGAN Overview.}
\textbf{(a) Training:} ViT-MAE predicts tokens for masked patches, supervised by a frozen VQGAN encoder.
\textbf{(b) Inference:} The model generates tokens from a masked input, which are decoded by VQGAN.}
  \label{fig:mae}
  \vspace{-1em}  
\end{figure}

\begin{figure*}[ht]
  \centering
  \vspace{-3mm} 
  \includegraphics[width=\linewidth]{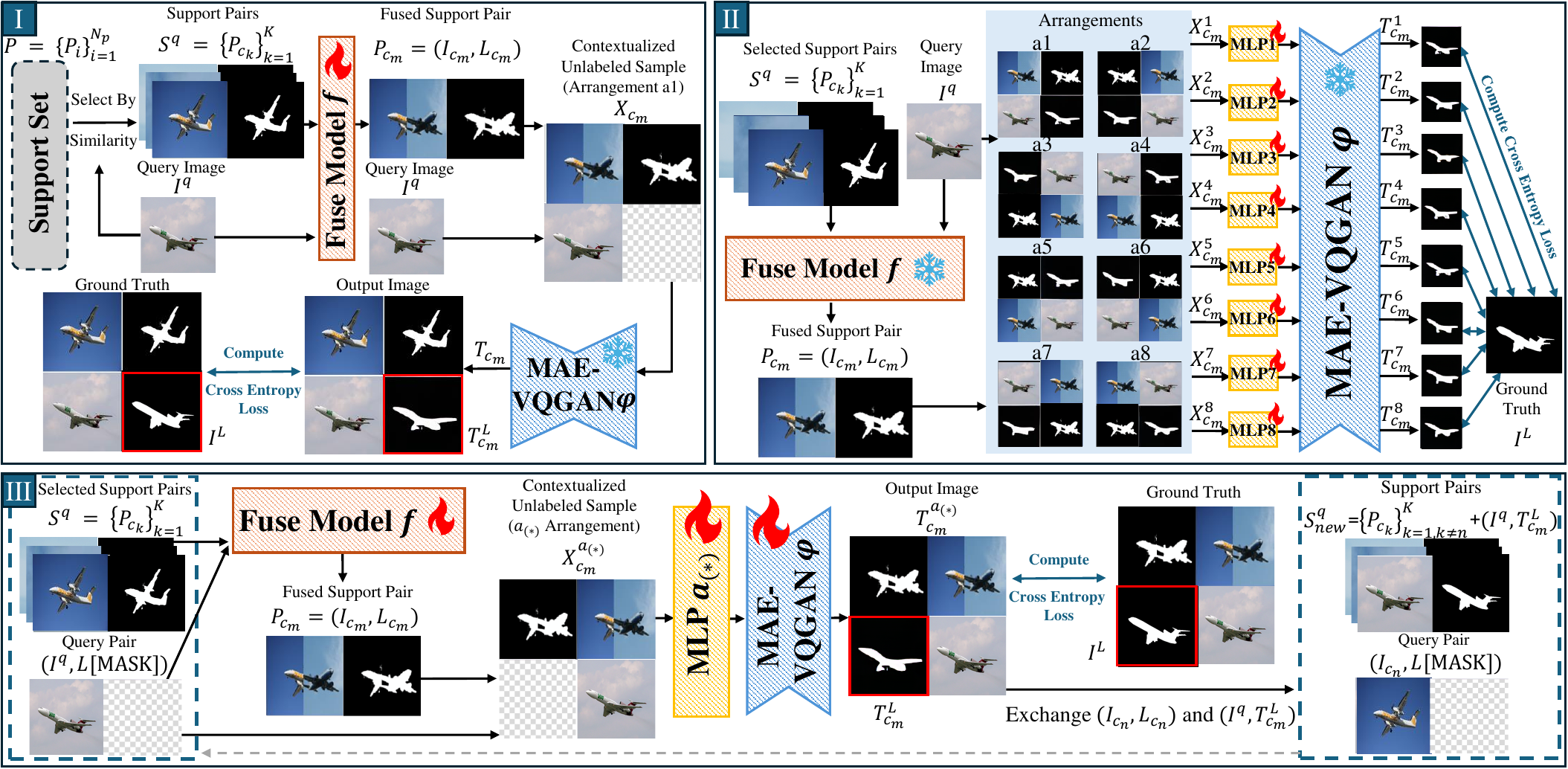} 

  \vspace{-1mm} 
  \caption{\textbf{Three-stage training pipeline.}
\textbf{(I) Preliminary Fusion Training:} The Fusion Module is first trained independently to establish a strong fusion capability.
\textbf{(II) Arrangement-Specific MLP Training:} Lightweight MLPs are trained for various geometric layouts to efficiently identify the optimal arrangements.
\textbf{(III) Joint Fine-tuning:} Finally, the whole model is fine-tuned on the optimal arrangements using a support–query swapping strategy to enhance overall performance and generalization.}
  \label{fig:pipeline}
  \vspace{-5mm} 
\end{figure*}

\section{Preliminaries}
\label{sec:pre}
\subsection{VICL Baseline}
Let the training set be \(Q = \{(I^q, L^q)\}_{q=1}^{N_q}\), where each query image \(I^q\) and its label \(L^q\) (e.g., bounding boxes or segmentation masks) lie in \(\mathbb{R}^{H_0/2 \times W_0/2 \times 3}\) with \(H_0 = W_0 = 224\). 
An auxiliary prompt database \(P = \{P_i = (I_i, L_i)\}_{i=1}^{N_p}\) provides task-specific contextual references.

For a given query \(I^q\), the top-\(K\) most similar prompts \(S^q = \{P_{c_k}\}_{k=1}^K\) are retrieved from \(P\), and the top-ranked one \(P_{c_1}\) is used to construct the contextualized sample:
\begin{equation}
X_{c_1} =
\begin{bmatrix}
I_{c_1} & L_{c_1} \\
I^q & \; L[\text{MASK}] \;
\end{bmatrix}
\in \mathbb{R}^{H_0 \times W_0 \times 3},
\end{equation}
where \(L[\text{MASK}]\) denotes the masked label region of the query image.
The constructed sample \(X_{c_1}\) is then input to an image inpainting model to reconstruct the missing \(L[\text{MASK}]\) region.

\subsection{MAE-VQGAN}
\label{sec:mae-vqgan}
For this experiment, we employ the MAE-VQGAN architecture~\cite{bar2022visual} for image inpainting, which has served as a foundational model in several subsequent works~\cite{sun2025exploring, xu2024towards, zhang2024instruct, zhang2023makes, wang2025embracing, xie2025test}. MAE-VQGAN combines a Vision Transformer (ViT)~\cite{vaswani2017attention, dosovitskiy2020image} encoder, trained with a Masked Autoencoder (MAE) objective~\cite{he2022masked}, and a Vector Quantised Generative Adversarial Network (VQGAN)~\cite{esser2021taming} decoder.

The model is trained to reconstruct occluded image patches by predicting their corresponding discrete tokens. During the training phase (Figure~\ref{fig:mae}a), the encoder's token predictions are optimized against the ground-truth VQGAN tokens via a cross-entropy loss. At inference (Figure~\ref{fig:mae}b), the ViT-MAE encoder generates a complete set of tokens, which the VQGAN decoder then synthesizes into the final output.

Given a constructed input \(X_{c_1}\), the MAE-VQGAN model, denoted as \(\varphi\), produces the completed image \(T_{c_1}\) as follows:
\begin{equation}
\label{eq:mae-vqgan}
T_{c_1} = \varphi(X_{c_1}).
\end{equation}


\section{Method}
\label{sec:method}
\subsection{Overview}

After pre-training the MAE-VQGAN (Section~\ref{sec:pre}) in line with previous approaches \cite{bar2022visual}. The overall pipeline is illustrated in Figure~\ref{fig:pipeline}. First, to ensure the model possesses sufficient fusion capability, we conduct preliminary training of the Fusion Module (Section~\ref{sec:fuse}). Building upon this, we freeze the core model and train a set of lightweight MLPs to each of the different spatial arrangements, so that we can rapidly identify the optimal layouts (Section~\ref{sec:arr}). Finally, we enhance the model’s generalization ability through a strategy of swapping support–query pairs, thereby achieving refined and improved overall performance (Section~\ref{sec:finetune}). We will elaborate on each stage in the following sections.

\subsection{Training of the Fusion Module}
\label{sec:fuse}

To distill the most relevant patterns and annotations from a set of support pairs into a single, high-quality fused representation, we first design a Fusion Module \textit{f}, which can provide each query sample with a targeted, fine-grained prompt. The training stage is illustrated in Figure~\ref{fig:pipeline}(I).

Given the feature representation of a query image $I^q$ and a support set 
$P = \{P_i=(I_i,L_i)\}_{i=1}^{N_p}$, 
we compute the cosine similarity between $I^q$ and the feature of each support image:
\begin{equation}
s_i = \cos\!\big(I^q, I_i\big), \quad i=1,\dots,N_p,
\end{equation}

The Top-$K$ most relevant pairs are selected according to $s_i$:

\begin{equation}
\{c_1, \dots, c_K\} \;=\; \operatorname{arg\,topK}_{i=1,\dots,N_p}\{s_i\},
\end{equation}

\begin{equation}
S^q = \{P_{c_k}\}_{k=1}^K,
\label{eq:K}
\end{equation}
where $\operatorname{arg\,topK}_{i=1,\dots,N_p}\{s_i\}$ denotes the indices of the $K$ support 
samples with the highest similarity scores.

We then input the selected support pair set $S^q$ together with the query image $I^q$ into the Fusion Module $\textit{f}$. 
The core of this model is a cross-attention mechanism that dynamically fuses information between the query and the selected supports. 
Formally, given the feature representations of the query and support images, the fused query feature is obtained as
\begin{equation}
{I}_{c_m}, \{\alpha_k\}_{k=1}^K = \mathrm{Attn}\!\big(I^q, \{I_{c_k}\}_{k=1}^K\big),
\end{equation}

\begin{equation}
    \alpha_k = \frac{\exp\!\big((I^q)^\top I_{c_k}\big)}{\sum_{j=1}^K \exp\!\big((I^q)^\top I_{c_j}\big)},
\end{equation}
where $\mathrm{Attn}(\cdot)$ denotes the cross-attention operation with the query feature $I^q$ as the Query and the support features $\{I_{c_k}\}$ as the Keys and Values. The $\{\alpha_k\}_{k=1}^K$ denotes the attention weights produced during the cross-attention.

To generate the fused target, we reuse the attention weights $\{\alpha_k\}$. 
Specifically, the fused target image $L_{c_m}$ is computed by the weighted sum of the support targets:
\begin{equation}
L_{c_m} = \sum_{k=1}^K \alpha_k \, L_{c_k}.
\end{equation}

The fused query feature ${I}_{c_m}$ and the fused target image $L_{c_m}$ together form a new fused support pair:
\begin{equation}
P_{c_m} = ({I}_{c_m}, L_{c_m}).
\end{equation}

Then we arrange the fused support pair $P_{c_m}=({I}_{c_m}, L_{c_m})$ and the query image $I^q$ into a $2 \times 2$ grid, forming a 
contextualized unlabeled sample $X_{c_m}$. 
Formally,
\begin{equation}
X_{c_m} = 
\begin{bmatrix}
{I}_{c_m} & L_{c_m} \\
I^q & \; L[\text{MASK}] \;
\end{bmatrix}.
\end{equation}

This construction defines an in-painting task for the frozen MAE-VQGAN $\varphi$: given the fused image, fused target, and query image, 
it must generate the missing target for the query. 
Specifically, we obtain the prediction as
\begin{equation}
T_{c_m} = \varphi(X_{c_m}) = \begin{bmatrix}
{I}_{c_m} & L_{c_m} \\
I^q & \; T^L_{c_m} \;
\end{bmatrix},
\end{equation}
where $T^L_{c_m}$ denotes the completed target image for $I^q$.

During this stage, the Fusion Module \textit{f} is optimized using two loss functions:

\paragraph{Alignment Loss.}  
To preserve semantic consistency between the query feature $I^q$ and the fused support feature $I_{c_m}$, 
we minimize their squared distance:
\begin{equation}
\mathcal{L}_{\text{align}} = \big\| I^q - I_{c_m} \big\|_2^2 .
\end{equation}

\paragraph{Cross-Entropy Loss.}  
As the main supervisory signal, we compute the cross-entropy loss between the predicted target feature $T^L_{c_m}$ 
and the ground-truth target feature $I^L$:
\begin{equation}
\mathcal{L}_{\text{CE}}= \text{CE}({T}^L_{c_m}, I^L) = - \sum_{j=1}^d I^L_j \log T^L_{c_m,j},
\end{equation}
where $d$ is the feature dimension.
 
The total training objective combines the two losses:
\begin{equation}
\mathcal{L}_{fuse} = \lambda \mathcal{L}_{\text{align}} + (1-\lambda) \mathcal{L}_{\text{CE}},
\label{eq:lamda}
\end{equation}
where $\lambda$ balances the alignment and cross-entropy terms.

By jointly optimizing these two loss functions, the Fusion Module \textit{f} learns to generate a high-quality, contextually-aligned fused representation for the query image based on highly relevant support samples, thereby laying a solid foundation for subsequent optimization stages.

\subsection{Decoupling Arrangement-Specific Priors}
\label{sec:arr}

To learn arrangement-specific knowledge and identify the optimal configurations, we introduce a search stage. To prevent interference from fusion and inpainting abilities, both the Fusion Module \textit{f} and the MAE-VQGAN $\varphi$ are kept frozen during this stage (Figure~\ref{fig:pipeline}(II)).

The fused support pair $P_{c_m}$ and query image $I^q$ are combined under predefined arrangements $\{a_i\}_{i=1}^8$ to form contextualized samples:
\begin{equation}
    X_{c_m}^i = \operatorname{Arrange}(P_{c_m}, I^q; a_i), \quad i=1,\dots,8.
\end{equation}

Instead of updating the backbone, we attach a lightweight, trainable module, denoted as $\text{MLP}_i$, to each arrangement. This module adopts an \textit{Adapter} architecture, which incorporates a residual connection around a feed-forward network (FFN). The computation is defined as:
\begin{equation}
    Z_{c_m}^i = X_{c_m}^i + \text{FFN}(X_{c_m}^i),
\end{equation}
where the FFN consists of two linear layers with a GELU activation function. It employs a bottleneck design to minimize trainable parameters, formulated as:
\begin{equation}
    \text{FFN}(X) = \boldsymbol{W}_2 \sigma(\boldsymbol{W}_1 X + \boldsymbol{b}_1) + \boldsymbol{b}_2,
\end{equation}
where $X$ is the input feature, $\boldsymbol{W}_1$ and $\boldsymbol{W}_2$ are the weight matrices, $\boldsymbol{b}_1$ and $\boldsymbol{b}_2$ are the bias terms, and $\sigma$ denotes the GELU activation.


The output of each module, $Z_{c_m}^i$, is then decoded by the frozen MAE-VQGAN $\varphi$ to produce a final prediction:
\begin{equation}
    T^i_{c_m} = \varphi(Z_{c_m}^i)= \begin{bmatrix}
{I}_{c_m} & L_{c_m} \\
I^q & \; T^{L,i}_{c_m} \;
\end{bmatrix}.
\end{equation}
The training objective for each \(\text{MLP}_i\) is to minimize the cross-entropy loss, defined as:
\begin{equation}
    \mathcal{L}^i_{arr} = \text{CE}(T^{L,i}_{c_m}, I^L).
\end{equation}

Following the convergence of all MLPs, we rank all candidate arrangements in descending order of their performance on a held-out test set. Let
$\{a_{(1)}, a_{(2)}, \dots, a_{(8)}\}$ denote the arrangements sorted such that
$\text{mIoU}_{(1)} \ge \text{mIoU}_{(2)} \ge \dots \ge \text{mIoU}_{(8)}$.
To reduce the computational cost of subsequent fine-tuning, we then select the top four arrangements as the \emph{preferred arrangement set}, $\mathcal{A}^*$:
\begin{equation}
    \mathcal{A}^* = \{a_{(*)}\} = \{ a_{(1)}, a_{(2)}, a_{(3)}, a_{(4)} \}.
\end{equation}

This lightweight strategy captures spatial priors with negligible parameter overhead. 
The MLPs specialize in modeling layout-specific geometric structure, while the remaining network components focus on semantic reconstruction, preserving the original architecture without costly retraining. 
The preferred arrangement set $\mathcal{A}^*$ is then used in the final joint fine-tuning stage.

\subsection{Bidirectional Joint Fine-tuning}
\label{sec:finetune}

To align the fusion capability of the Fusion Module, the spatial reasoning of the MLPs, and the inpainting ability of the MAE-VQGAN, we perform end-to-end joint fine-tuning (Figure~\ref{fig:pipeline} (III)). 
Using the $\mathcal{A}^*$, we construct the contextualized sample $X_{c_m}^{a_{(*)}}$, which is processed by the MLP$_{a_{(*)}}$ and MAE-VQGAN $\varphi$ to produce the output
\begin{equation}
T_{c_m}^{a_{(*)}} = \varphi(\text{MLP}_{a_{(*)}}(X_{c_m}^{a_{(*)}})).
\end{equation}

The model is optimized by the cross-entropy loss
\begin{equation}
\mathcal{L}_{finetune} = \text{CE}(T_{c_m}^{a_{(*)}}, I^L),
\end{equation}
allowing simultaneous fine-tuning of \textit{f} and MLP$_{a_{(*)}}$ to enhance fusion capability and generalization.

To enhance the robustness and prevent overfitting, we further introduce a bidirectional fine-tuning mechanism. Motivated by the principle of informational symmetry \cite{cover1999elements}: \textit{a robust fusion model should not merely learn a one-way mapping from support to query. Instead, the fusion process should be reversible.} This symmetry discourages shortcut learning and promotes capturing shared structural relations. 

Based on this, we introduce an online data augmentation strategy. 
After obtaining the prediction $T_{c_m}^{L}$ for the current query $I^q$, we form a new pair:
\begin{equation}
P_\text{new} = (I^q, T_{c_m}^{L}).
\end{equation}

From the Top-$K$ support set $S^q = \{P_{c_k}\}_{k=1}^K$, the Top-$N$ most similar pairs are selected:
\begin{equation}
Q^S = \{P_{c_n}\}_{n=1}^N, \quad N \le K, \quad P_{c_n} = (I_{c_n}, L_{c_n}).
\label{eq:N}
\end{equation}

During each training iteration, we perform $N$ sub-iterations, each replacing one $P_{c_n}$ in $Q^S$ with $P_\text{new}$:
\begin{equation}
S^q_\text{new} = \big( S^q \setminus \{P_{c_n}\} \big) \cup \{ P_\text{new} \}, \quad n = 1,\dots,N,
\end{equation}
plus one additional sub-iteration without replacement, resulting in $N+1$ sub-iterations. 
In each sub-iteration, $S^q_\text{new}$ serves as the support and $P_{c_n}$ as the query. The bidirectional joint optimization substantially strengthens the generalization and stability of the learned fusion logic.
\section{Experiments}
\label{sec:formatting}
\begin{table*}
\centering
\small
\begin{tabular}{p{1.1cm}lccccccc}
\toprule
\textbf{Type} & \textbf{Model} & \multicolumn{5}{c}{\textbf{Seg. (mIoU $\uparrow$)}} & \textbf{Det. (mIoU $\uparrow$)} & \textbf{Col. (MSE $\downarrow$)} \\
\cmidrule(lr){3-7}
& & \textbf{Fold-0} & \textbf{Fold-1} & \textbf{Fold-2} & \textbf{Fold-3} & \textbf{Mean} & & \\
\midrule\midrule
\multirow{5}{*}{\shortstack[l]{Single\\Prompt\\Selection}} 
& Random \cite{bar2022visual} & 28.66 & 30.21 & 27.81 & 23.55 & 27.56 & 25.45 & 0.67 \\
& UnsupPR \cite{zhang2023makes} & 34.75 & 35.92 & 32.41 & 31.16 & 33.56 & 26.84 & 0.63 \\
& SupPR \cite{zhang2023makes} & 37.08 & 38.43 & 34.40 & 32.32 & 35.56 & 28.22 & 0.63 \\
& Prompt-Self \cite{sun2025exploring} & 35.69 & 38.25 & 35.86 & 33.37 & 35.79 & 28.08 & 0.63 \\
& Partial2Global \cite{xu2024towards} & 38.81 & 41.54 & 37.25 & 36.01 & 38.40 & 30.66 & 0.58 \\
& InMeMo \cite{zhang2024instruct} & 41.65 & 47.68 & 42.43 & 40.80 & 43.14 & 43.21 & - \\
& Task-Level Prompting \cite{zhu2025exploring} & 39.09{\tiny\,±0.77} & 44.37{\tiny\,±0.98} & 37.93{\tiny\,±0.42} & 32.40{\tiny\,±1.06} & 38.45 & 29.03{\tiny\,±2.84} & 0.62{\tiny\,±1.27} \\
\midrule
\multirow{2}{*}{Voting} 
& Prompt-Self$_{\text{w/ voting}}$ \cite{sun2025exploring} & 42.48 & 43.34 & 39.76 & 38.50 & 41.02 & 29.83 & - \\
& Partial2Global$_{\text{w/ voting}}$ \cite{xu2024towards} & 43.23 & 45.50 & 41.79 & 40.22 & 42.69 & 32.52 & - \\
\midrule
\multirow{2}{*}{\cellcolor{white}Fusion} 
& CONDENSER$_{K=1}$ \cite{wang2025embracing} & 42.13 & 50.31 & 42.20 & 41.90 & 44.14 & 43.22 & 0.56 \\
& CONDENSER$_{K=16}$ \cite{wang2025embracing} & 45.53 & 52.06 & 44.33 & 44.58 & 46.63 & 44.64 & \underline{0.54} \\
\midrule
\multirow{2}{*}{\shortstack[l]{Ensemble}} 
& \cellcolor{lightblue}Ours$_{K=1}$ 
& \cellcolor{lightblue}\underline{48.57{\tiny\,±0.06}}
& \cellcolor{lightblue}\underline{54.26{\tiny\,±0.09}}
& \cellcolor{lightblue}\underline{47.23{\tiny\,±0.07}}
& \cellcolor{lightblue}\underline{45.74{\tiny\,±0.10}}
& \cellcolor{lightblue}\underline{48.95}
& \cellcolor{lightblue}\underline{44.19}{\tiny\,±0.08} 
& \cellcolor{lightblue}{0.56}{\tiny\,±0.29} \\
& \cellcolor{lightpurple}Ours$_{K=16}$ 
& \cellcolor{lightpurple}\textbf{49.32{\tiny\,±0.06}}
& \cellcolor{lightpurple}\textbf{55.42{\tiny\,±0.09}}
& \cellcolor{lightpurple}\textbf{47.57{\tiny\,±0.07}}
& \cellcolor{lightpurple}\textbf{48.98{\tiny\,±0.10}}
& \cellcolor{lightpurple}\textbf{50.32}
& \cellcolor{lightpurple}\textbf{45.07}{\tiny\,±0.07} 
& \cellcolor{lightpurple}\textbf{0.53}{\tiny\,±0.26} \\
\bottomrule
\end{tabular}
\caption{Performance comparison across different tasks. \textbf{Bold} indicates the best performance, and \underline{underlined} indicates the second-best.}
\label{tab:main}
\end{table*}

\begin{figure*}[!t]
\centering
\begin{subfigure}[t]{0.245\textwidth}
    \includegraphics[width=\linewidth]{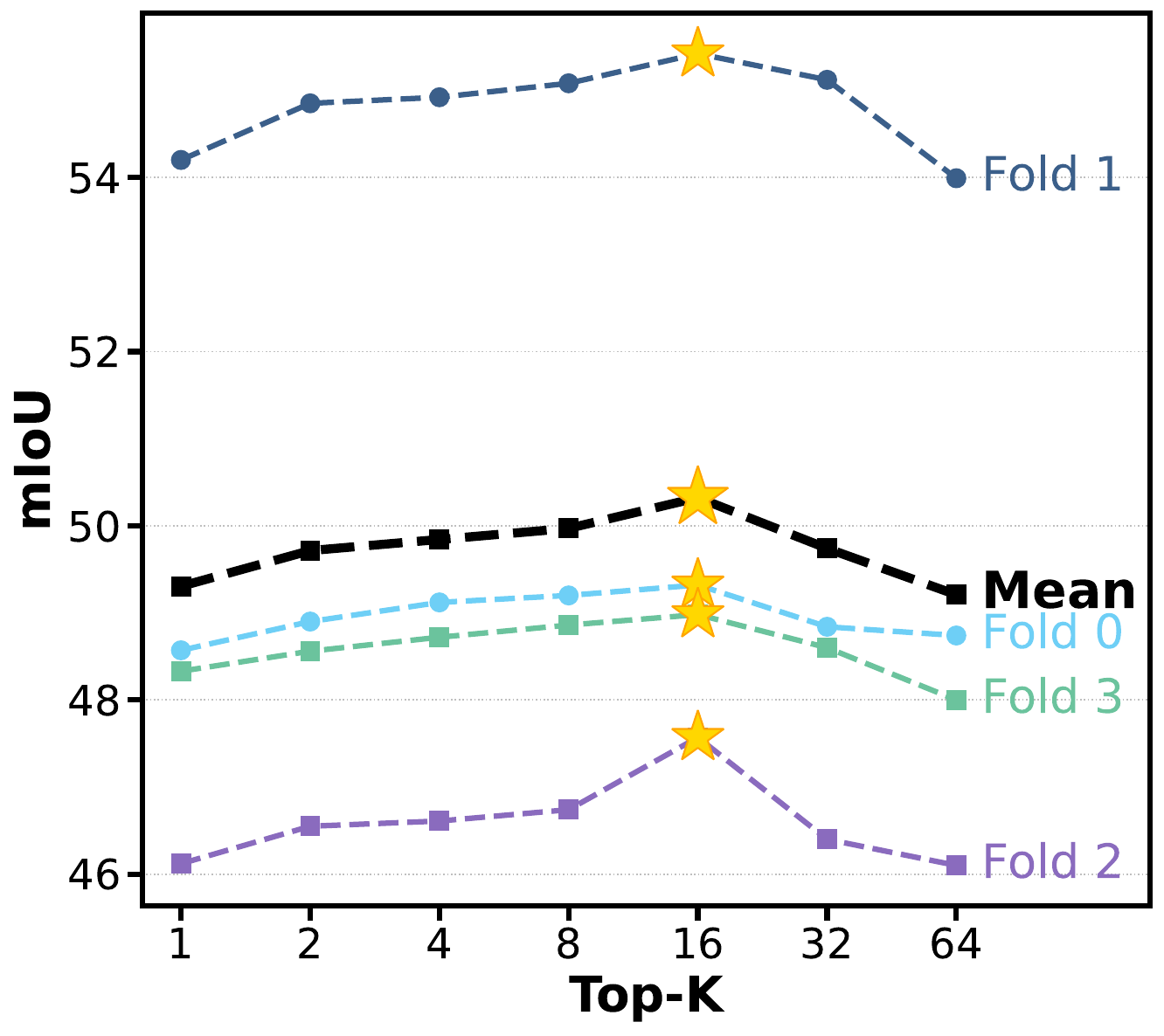}
    \caption{Impact of $K$ in Fusion Module.}
    \label{subfig1}
\end{subfigure}
\hfill
\begin{subfigure}[t]{0.245\textwidth}
    \includegraphics[width=\linewidth]{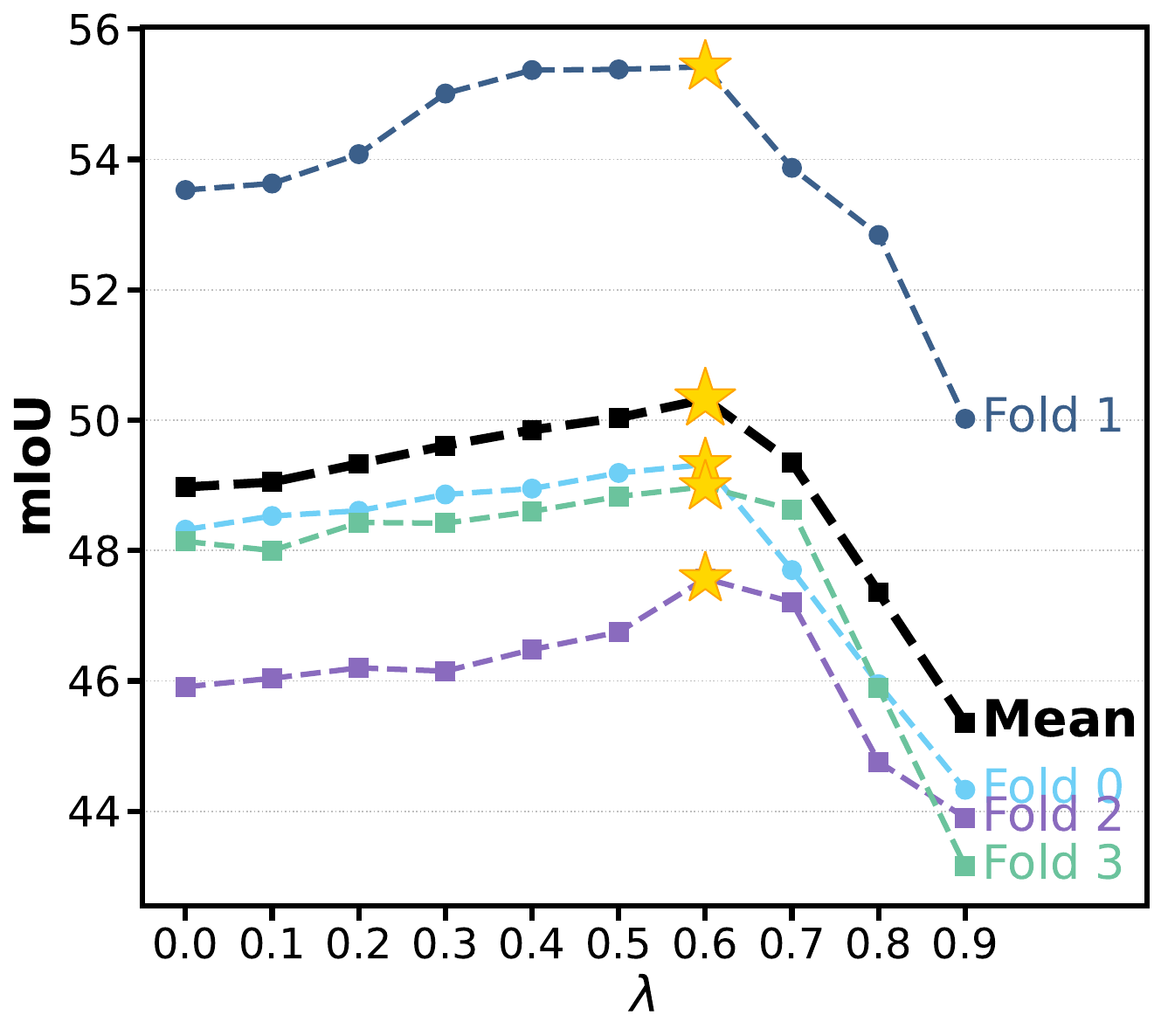}
    \caption{Effect of the loss balancing $\lambda$.}
    \label{subfig2}
\end{subfigure}
\hfill
\begin{subfigure}[t]{0.245\textwidth}
    \includegraphics[width=\linewidth]{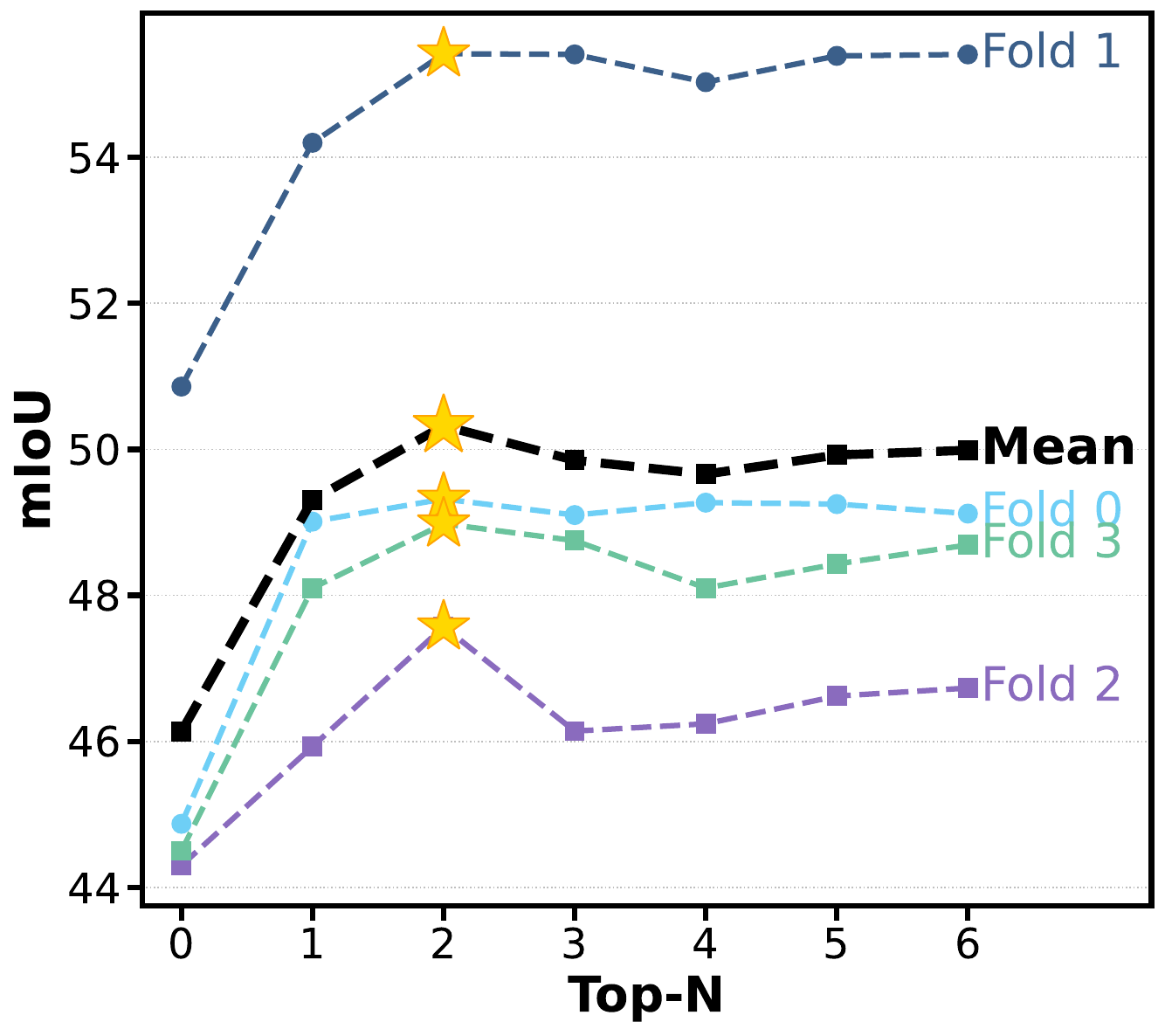}
    \caption{Impact of $N$ for joint fine-tuning.}
    \label{subfig3}
\end{subfigure}
\hfill
\begin{subfigure}[t]{0.245\textwidth}
    \includegraphics[width=\linewidth]{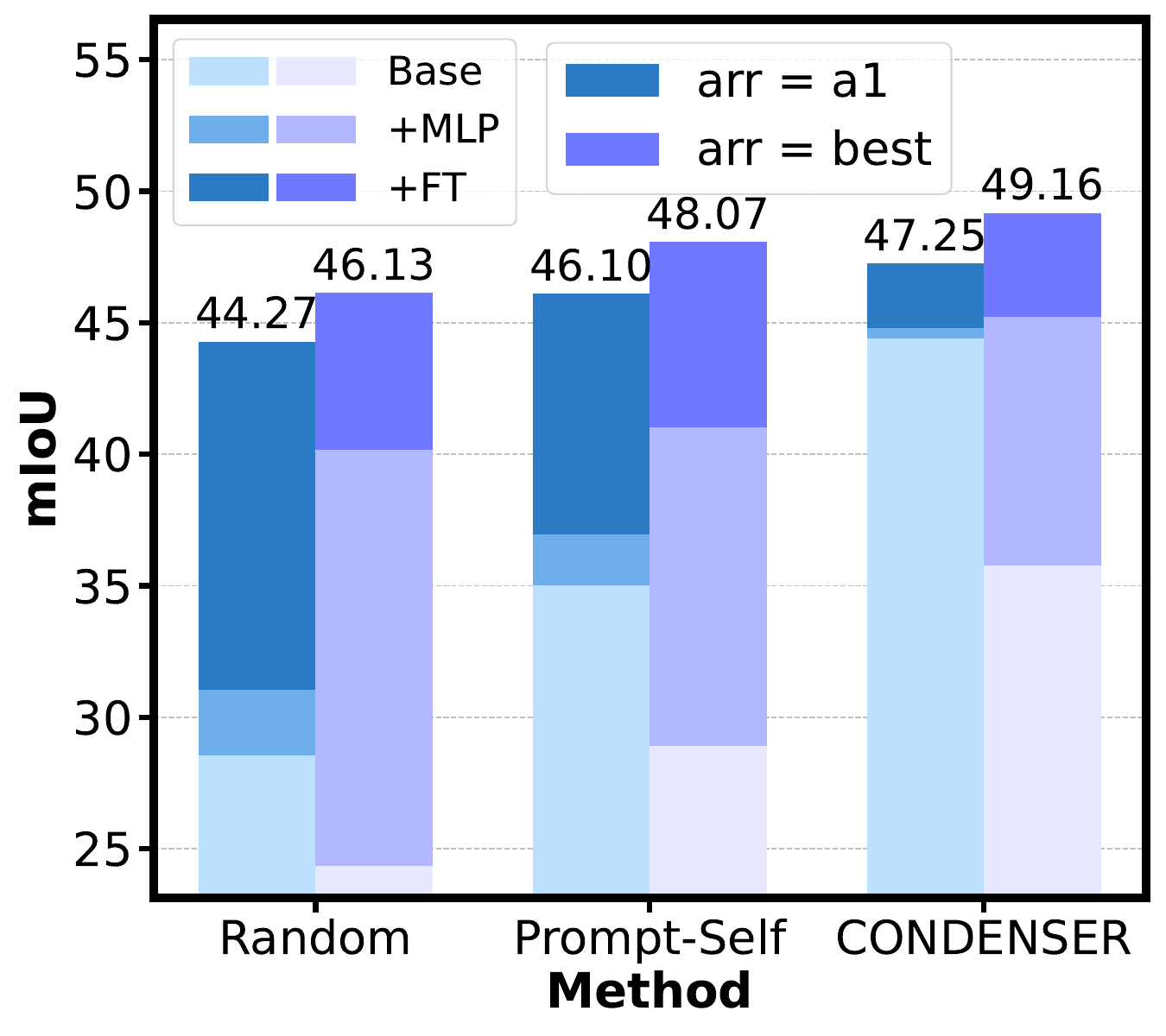}
    \caption{Generalization across methods.}
    \label{subfig4}
\end{subfigure}
\caption{Ablation studies on key hyperparameters and module generalization.}
\label{fig:ablations}
\end{figure*}

\subsection{Datasets and Implementation Details}

We evaluate our framework on three public datasets, PASCAL-5$^i$ \cite{shaban2017one}, PASCAL VOC 2012 \cite{everingham2015pascal}, and ImageNet-1K \cite{russakovsky2015imagenet}, corresponding to foreground segmentation, single object detection, and image colorization, respectively. Further details regarding the datasets and implementation are deferred to the supplementary material.

\subsection{Baseline}

We comprehensively compare Our method with representative baselines categorized into three groups: 
(i) \textbf{Single Prompt Selection}: including random selection~\cite{bar2022visual}, VPR~\cite{zhang2023makes} with optimized prompt retrievers, Prompt-SelF \cite{sun2025exploring} with direct prompt ranking, Partial2Global~\cite{xu2024towards} with a hierarchical prompt ranker, InMeMo~\cite{zhang2024instruct} with learnable visual prompts, and task-level prompting~\cite{zhu2025exploring} with task-specific prompt discovery; 
(ii) \textbf{Voting}: Prompt-SelF~\cite{sun2025exploring} and Partial2Global~\cite{xu2024towards}, which aggregate predictions from different arrangements of the same query–prompt pairs on the input canvas; 
(iii) \textbf{Prompt Fusion}: \textsc{CONDENSER} \cite{wang2025embracing}, an external plugin that integrates fine-grained contextual information across multiple prompts.


\subsection{Overall Performance}

As shown in Table \ref{tab:main}, our proposed method consistently surpasses existing approaches across segmentation, detection, and colorization tasks. Moreover, following the settings of InMeMo \cite{zhang2024instruct}, Prompt-SelF \cite{sun2025exploring} and CONDENSER \cite{wang2025embracing}, we construct a COCO-5$^i$ dataset from MSCOCO \cite{lin2014microsoft} that shares the same category splits as Pascal-5$^i$. We train our model on COCO-5$^i$ and evaluate it on Pascal-5$^i$. As presented in Table \ref{tab:coco}, our approach also demonstrates superior generalization performance. These results highlight the effectiveness of our end-to-end ensemble strategy that integrates both prompt fusion and prompt voting mechanisms.

\subsection{Ablation of Structure}

To assess the contribution of each component, we perform an ablation study summarized in Table~\ref{tab:structure}. Variants (1)–(4) are derived from the full model (5) by removing or modifying specific modules. Replacing the Fusion Module with simple averaging (1) reduces performance, indicating that prompt fusion helps integrate contextual cues. Removing attention reuse and using plain averaging (2) also leads to a drop, showing that weight of raw images is informative for target images. Notably, removing the residual connection in the MLP (3) results in a significant performance decrease, while discarding the bottleneck structure (4) further weakens representation. This highlights that the residual connection is crucial for preserving original feature information, while the 2-layer bottleneck structure not only reduces trainable parameters but also distills a more compact and effective representation. Overall, these results validate the necessity of our architectural design.

\subsection{Hyperparameter Sensitivity and Module Extensibility}

First, we investigate the influence of the number of support pairs, $K$, used for fusion, as defined in Equation~\eqref{eq:K}.
As shown in Figure~\ref{subfig1}, we vary $K$ from 1 to 64. The performance, measured in mIoU, generally improves as more support pairs are included, reaching a peak at $K=16$.
Beyond this point, performance slightly declines, likely due to the introduction of less relevant or noisy information from lower-ranked support pairs.

Next, we analyze the sensitivity of our model to the hyperparameter $\lambda$, which balances the alignment loss and the cross-entropy loss in our total training objective, as shown in Equation~\eqref{eq:lamda}.
Figure~\ref{subfig2} illustrates the model's performance as $\lambda$ is varied from 0.0 to 0.9. This trend indicates that the alignment loss enhances semantic consistency up to the peak at $\lambda = 0.6$. Beyond this point, excessive emphasis on alignment disrupts the balance with cross-entropy optimization, leading to performance degradation.

We also explore the optimal number of pairs, $N$, to select from the support set for our iterative joint fine-tuning strategy, as described in Equation~\eqref{eq:N}.
This hyperparameter determines how many of the most similar support samples are iteratively swapped with the query for adaptation.
As seen in Figure~\ref{subfig3}, the performance steadily increases and peaks at $N=2$. This suggests that N=2 is the most effective choice, as it represents the minimum value required to achieve peak performance, striking the optimal trade-off between accuracy and computational efficiency.

Finally, to validate the general utility of our MLP module and joint fine-tuning strategy (+FT), we integrated them into several baseline methods. As shown in Figure~\ref{subfig4}, both consistently improve performance across all baselines, demonstrating their effectiveness as plug-and-play components. Notably, the optimal arrangement (\textit{arr = best}) initially underperforms the sequential layout (\textit{arr = a1}) when models are trained only on \textit{a1}, indicating poor generalization to better layouts. After incorporating our arrangement-aware module, however, \textit{best} achieves an mIoU of 49.16, surpassing \textit{a1} (47.25). This reversal demonstrates that our framework effectively captures geometric prompt arrangement, unlocking the full potential of optimal prompt selection.

\begin{table}
\centering
\footnotesize
\setlength{\tabcolsep}{3pt}
\begin{tabular}{p{3cm}p{0.78cm}p{0.78cm}p{0.78cm}p{0.78cm}p{0.78cm}}
\toprule
\textbf{Model} & \multicolumn{5}{c}{\textbf{Seg. (mIoU $\uparrow$)}} \\
\cmidrule(lr){2-6}
& \textbf{Fold-0} & \textbf{Fold-1} & \textbf{Fold-2} & \textbf{Fold-3} & \textbf{Mean} \\
\midrule\midrule
InMeMo \cite{zhang2024instruct} & 38.74 & 43.82 & 40.45 & 37.12 & 40.03 \\
Prompt-SelF \cite{sun2025exploring} & 40.13 & 42.14 & 37.84 & \underline{38.52} & 39.66 \\
CONDENSER$_{K=1}$ \cite{wang2025embracing} & 40.39 & 44.54 & 40.23 & 36.33 & 40.37 \\
CONDENSER$_{K=16}$ \cite{wang2025embracing} & 40.37 & 44.85 & 41.03 & 35.84 & 40.52 \\
\rowcolor{lightblue}
Ours$_{K=1}$ & \textbf{42.86} & \underline{47.05} & \textbf{42.17} & 38.02 & \underline{42.53} \\
\rowcolor{lightpurple}
Ours$_{K=16}$ & \underline{42.56} & \textbf{47.18} & \underline{41.92} & \textbf{38.71} & \textbf{42.59} \\
\bottomrule
\end{tabular}
\caption{Cross-dataset performance evaluation. We train models on COCO-5$^i$ and test on Pascal-5$^i$ for the segmentation task.}
\label{tab:coco}
\end{table}

\begin{table}
\centering
\footnotesize
\setlength{\tabcolsep}{5.5pt}
\begin{tabular}{clccccc} 
\toprule
\textbf{ID} & \textbf{Setting} & \multicolumn{5}{c}{\textbf{Seg. (mIoU $\uparrow$)}}\\
\cmidrule(lr){3-7}
&  & \textbf{Fold-0} & \textbf{Fold-1} & \textbf{Fold-2} & \textbf{Fold-3} & \textbf{Mean} \\ 
\midrule\midrule
(1) & w/o Fusion      & \underline{49.07} & 53.62 & \underline{47.56} & 47.08 & 49.33 \\
(2) & w/o Reuse       & 48.27 & 54.35 & 47.54 & \underline{47.86} & \underline{49.51} \\
(3) & w/o Residual    & 31.54 & 45.19 & 39.44 & 33.46 & 37.41 \\
(4) & 1-layer MLP     & 46.95 & \underline{55.01} & 47.29 & 47.50 & 49.19 \\
\rowcolor{lightpurple}
(5) & Full Model      & \textbf{49.32} & \textbf{55.42} & \textbf{47.57} & \textbf{48.98} & \textbf{50.32} \\
\bottomrule
\end{tabular}
\caption{Ablation study of model components.}
\label{tab:structure}
\end{table}

\begin{table*}[ht]
\centering
\small
\setlength{\tabcolsep}{7.68pt}
\begin{tabular*}{0.98\textwidth}{@{\extracolsep{\fill}}cp{1.2cm}ccccccc}
\toprule
\textbf{ID} & \textbf{Model} & \multicolumn{2}{c}{\textbf{Model Metrics}} & \textbf{Setting} & \multicolumn{4}{c}{\textbf{System Performance}} \\
\cmidrule(lr){3-4} \cmidrule(lr){6-9}
& & \textbf{Params (M)} & \textbf{GFLOPs} & & \textbf{Train Time} & \textbf{Peak Mem. (GB)} & \textbf{Infer Time (ms)} & \textbf{FPS} \\
\midrule\midrule
(1) & FUSE & 52.07 & 13.265 & w/o\hspace{0.6em}MLP & 52m36.86s & 17.95 & 256.39{\tiny\,±5.66} & 31.20 \\
\rowcolor{lightpurple}
(2) & MLP & 4.01 & 0.0021 & w/\hspace{1.1em}MLP& 52m47.91s & 17.38 & 256.90{\tiny\,±5.10} & 31.14 \\
\bottomrule
\end{tabular*}

\vspace{0.4em}

\begin{tabular*}{\textwidth}{@{\extracolsep{\fill}}clccccccccc}
\toprule
\textbf{ID} & \textbf{Setting} & \textbf{a1} & \textbf{a2} & \textbf{a3} & \textbf{a4} & \textbf{a5} & \textbf{a6} & \textbf{a7} & \textbf{a8} & \textbf{Mean} \\
\midrule\midrule
(3) & Base & 35.02{\tiny,±0.07} & 11.04{\tiny,±0.13} & 14.60{\tiny,±0.15} & 28.92{\tiny,±0.09} & 24.85{\tiny,±0.11} & 12.83{\tiny,±0.15} & 12.36{\tiny,±0.15} & 27.17{\tiny,±0.10} & 20.85{\tiny,±0.12} \\
(4) & Base-FT & 46.05{\tiny,±0.06} & 47.07{\tiny,±0.06} & 47.93{\tiny,±0.05} & 47.34{\tiny,±0.06} & 48.03{\tiny,±0.05} & 48.49{\tiny,±0.05} & \textbf{48.67}{\tiny,±0.05} & 48.67{\tiny,±0.05} & 47.78{\tiny,±0.05} \\
\midrule
(5) & Fuse & 44.42{\tiny,±0.08} & 44.17{\tiny,±0.08} & 43.74{\tiny,±0.09} & 44.69{\tiny,±0.07} & 44.56{\tiny,±0.07} & 45.39{\tiny,±0.06} & 43.88{\tiny,±0.08} & 43.89{\tiny,±0.08} & 44.34{\tiny,±0.08} \\
(6) & Fuse-FT & 47.25{\tiny,±0.06} & 47.59{\tiny,±0.05} & 47.30{\tiny,±0.06} & 48.97{\tiny,±0.05} & 47.43{\tiny,±0.06} & \textbf{49.14}{\tiny,±0.05} & 48.24{\tiny,±0.05} & 48.27{\tiny,±0.05} & 48.02{\tiny,±0.05} \\
\midrule
\rowcolor{lightblue}
(7) & MLP & 44.12{\tiny,±0.07} & 44.38{\tiny,±0.08} & 44.81{\tiny,±0.07} & 45.24{\tiny,±0.07} & 44.88{\tiny,±0.07} & 45.26{\tiny,±0.06} & 43.98{\tiny,±0.09} & 44.54{\tiny,±0.08} & 44.65{\tiny,±0.08} \\
\rowcolor{lightpurple}
(8) & MLP-FT & \textbf{47.26}{\tiny,±0.06} & \textbf{47.73}{\tiny,±0.05} & \textbf{48.85}{\tiny,±0.05} & \textbf{49.32}{\tiny,±0.05} & \textbf{49.20}{\tiny,±0.05} & 48.50{\tiny,±0.05} & 47.95{\tiny,±0.05} & \textbf{48.70}{\tiny,±0.05} & \textbf{48.44}{\tiny,±0.05} \\
\bottomrule
\end{tabular*}
\caption{Model efficiency and performance comparison. The results validate and demonstrate the lightweight design of MLP and the effectiveness of the Synergistic Training strategy. -FT denotes joint fine-tuning.}
\label{tab:ABLATION}
\end{table*}

\begin{figure*}
  \centering
  \vspace{-3mm} 
  \includegraphics[width=\linewidth]{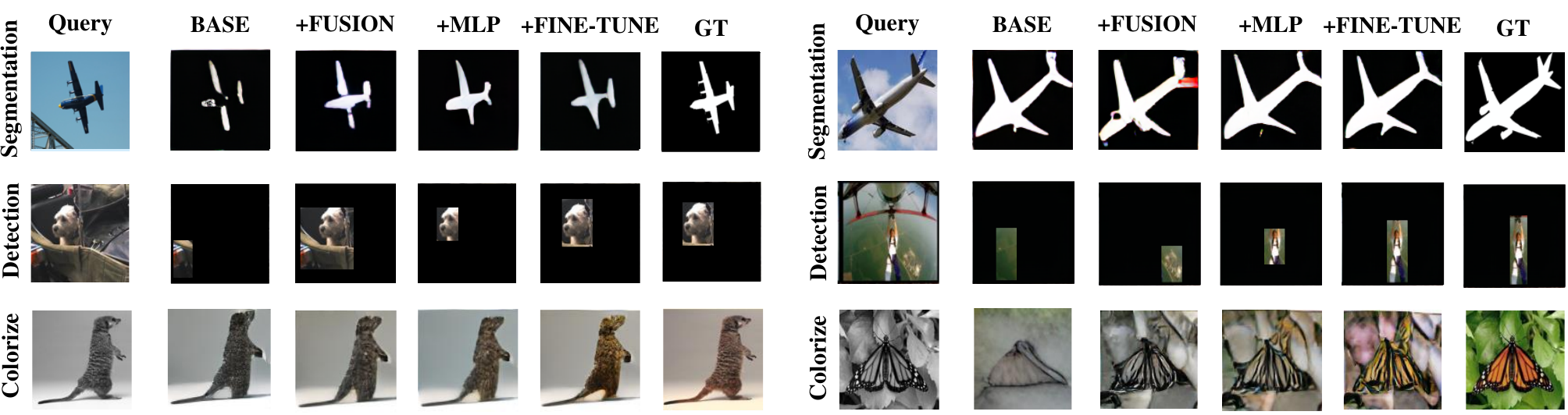} 

  \vspace{-1mm} 
  \caption{Visualization across tasks, showing progressive improvements from adding the Fusion Module, the MLP, and joint fine-tuning.}
  \label{fig:visual}
  \vspace{-5mm} 
\end{figure*}

\subsection{Analysis of Component Effectiveness}

To verify that the introduced MLP module is lightweight, we compare the Fusion Module settings in Table \ref{tab:ABLATION}. As shown in (1) and (2), our MLP module contains only 7.7\% of the parameters and less than 0.02\% of the computational cost in GFLOPs compared to Fusion Module. Moreover, introducing this module only slightly affects training time, peak memory, inference latency, and FPS. This evidence strongly supports the conclusion that the MLP adds valuable spatial layout awareness at a minimal computational expense. Lightweight computer vision models are particularly valuable for deployment on mobile and edge devices where computational resources are often limited. 

Base (3) performs noticeably worse than the other settings, indicating that explicitly modeling layout-specific characteristics is essential. This further supports our argument against the conclusions drawn in \cite{sun2025exploring} that a single spatially “blind” frozen model fundamentally limits performance. Additionally, No matter finetune or not, MLP (7)(8) outperforms Fuse (5)(6) and Base (3)(4). This confirms our design: fine-tuning a small layout-specific MLP is more effective than fine-tuning the entire model, as it cleanly separates spatial encoding from semantic processing. This separation of concerns allows for more targeted and efficient learning, leading to improved overall performance. Finally, comparing before and after fine-tuning across all settings shows consistent performance gains, demonstrating that our Synergistic Training strategy jointly improves contextual fusion and reconstruction. By forcing the model to work in both a forward and reverse direction, this bidirectional fine-tuning substantially strengthens the generalization and stability of the learned fusion logic.

\subsection{Qualitative Analysis and Generalization}
Figure \ref{fig:visual} presents qualitative results highlighting the progressive improvements of our framework. In the segmentation task, the baseline MAE-VQGAN produces fragmented masks, while the Fusion Module enhances contextual integration but still leaves some structural errors. Incorporating the MLP, which explicitly models the spatial layout of support pairs, further refines the outputs, yielding masks with improved structural integrity. Finally, our synergistic fine-tuning produces highly accurate masks with sharp boundaries, demonstrating the strong interplay between arrangement-aware MLPs and joint optimization. Beyond segmentation, the framework delivers high-fidelity results on tasks such as object detection and image coloring, showcasing its versatility.

\section{Conclusions}
\label{sec:formatting}

In this work, we propose a new end-to-end framework for VICL that overcomes the limitations of relying on a single prompt and overlooking the potential of spatial arrangement. Our approach integrates an adaptive Fusion Module to combine complementary information from multiple prompts, and introduces lightweight arrangement-specific MLPs to explicitly encode spatial layouts. Additionally, our query–support swapping strategy enables synergistic training between fusion and reconstruction. Experiments on segmentation, detection, and colorization demonstrate the robustness and superior performance of our method. This work advances VICL toward a more holistic contextual understanding, and opens promising directions for supporting arbitrary arrangements and more complex generative tasks.

\bibliographystyle{plain}
\newpage
\bibliography{reference}

\clearpage        
\appendix         
\section*{Appendix} 

\crefname{section}{Sec.}{Secs.}
\Crefname{section}{Section}{Sections}
\Crefname{table}{Table}{Tables}
\crefname{table}{Tab.}{Tabs.}

\def\cvprPaperID{***} 
\def\confName{CVPR}
\def\confYear{2026}


\section{Overview}
\label{sec:over}
We present additional details regarding related works, datasets, implementation, experiments, qualitative results, and algorithmic specifics in this Supplementary Material. The Supplementary Material is organized as follows:
\begin{itemize}
    \item Section~\ref{sec:rela} provides a comprehensive review of related works.
    \item Section~\ref{sec:data} offers detailed information on the datasets used and the specifics of our implementation.
    \item Section~\ref{sec:ex} presents further details on our experimental configurations.
    \item Section~\ref{sec:qu} provides qualitative results to visually demonstrate the performance and versatility of our proposed model.
    \item Section~\ref{sec:code} presents the pseudo-code for our method.
\end{itemize}

\section{Related Works}
\label{sec:rela}

\subsection{In-Context Learning}
ICL is first introduced by GPT-3 \cite{brown2020language}, reformulates NLP tasks as prompt-based text completion. Without parameter updates, LLMs can quickly adapt to new reasoning tasks or unseen patterns by imitating a few in-context prompts, effectively serving as few-shot \cite{lu2021fantastically,wu2022self,wei2022chain}. Its success has extended to multi-modal domains. For instance, Flamingo \cite{alayrac2022flamingo} expands LLM inputs to images and videos while keeping language as the interface, enabling tasks like image captioning and visual question answering through mixed text-visual prompts \cite{achiam2023gpt,team2023gemini,team2024gemini}.

\subsection{Visual In-Context Learning}
Inspired by the success of ICL in language and multi-modal domains, recent research extends this paradigm to the pure vision field VICL, which aims to build a general vision model that performs arbitrary tasks at inference using only a few visual prompts. Pioneering works such as MAE-VQGAN \cite{bar2022visual}, Painter \cite{wang2023images}, and SegGPT \cite{wang2023seggpt} unify diverse visual tasks as inpainting or masked image modeling, showing strong potential. Building on this, follow-up studies explore how to improve VICL, especially in visual prompt selection \cite{sun2025exploring,balazevic2023towards}. prompts include VPR \cite{zhang2023makes} (reducing the resolution of sub-images to meet input requirements), InMeMo \cite{zhang2024instruct} (prompt tuning to reduce reliance on high-quality prompts), and Partial2Global \cite{xu2024towards} (ranking models with global supervision to address contrastive learning limits).

Despite recent progress, most studies still focus on single-prompt learning, and the exploration of multi-prompt and arrangement-based strategies remains limited. Although CONDENSER \cite{wang2025embracing} pioneers to explore the benefits of fusing multiple prompts, their modular design separates fusion from inpainting, leaving the generator as a passive receiver. In contrast, our end-to-end framework jointly optimizes fusion and generation, and further incorporates arrangement-aware MLPs, leading to a more coherent and powerful representation than sequential fusion approaches.

\section{Datasets and Implementation Details}
\label{sec:data}

\subsection{Evaluation Metrics}
Our model's performance is evaluated using standard, task-specific metrics to ensure a fair comparison with prior work. For tasks requiring precise spatial localization, namely foreground segmentation and single-object detection, we report the \textbf{mean Intersection-over-Union (mIoU)}. This metric effectively quantifies the accuracy of the predicted masks and bounding boxes. For the generative task of image colorization, which can be framed as a regression problem at the pixel level, we measure the performance using the \textbf{Mean Squared Error (MSE)} to assess the difference between the predicted color values and the original ground-truth image.

\subsection{Datasets}

Our experimental pipeline is built upon three widely-used public datasets: PASCAL-5$^i$, PASCAL VOC 2012, and ImageNet-1K. We process these datasets to specifically adapt them for three downstream tasks: foreground segmentation, single object detection, and image colorization.

\subsubsection{PASCAL-5$^i$} This dataset~\cite{shaban2017one} is a variant of PASCAL VOC 2012, specifically designed for few-shot segmentation. It organizes 20 object categories into 4 disjoint folds, each containing 5 classes. We leverage this structure to evaluate our model's \textbf{foreground segmentation} performance. We adopt the training sample configuration proposed by~\cite{zhang2024instruct}, using 2,286, 3,425, 5,883, and 2,086 samples for the four respective folds. By training and testing on these splits, we can effectively measure the model's segmentation performance across various class combinations, reporting the mean mIoU as the final metric.

\subsubsection{PASCAL VOC 2012.} As a classic benchmark in object detection, we utilize this dataset~\cite{everingham2015pascal} to assess the model's proficiency in \textbf{single object detection}. For experimental clarity, we filter the dataset to retain only images that contain a single annotated object. The model is trained using 612 curated in-context samples. The model's predictions (i.e., bounding boxes) are compared against the ground-truth annotations provided by the dataset, and accuracy is determined by the mean mIoU score.

\subsubsection{ImageNet-1K (ILSVRC2012).} 

We explore the potential of this large-scale image classification dataset~\cite{russakovsky2015imagenet} by applying it to the task of \textbf{image colorization}. We devise a specific training pipeline: first, we randomly sample a 50k-image subset from the full 1.2M training set to serve as our data source. Subsequently, these images are converted from color to grayscale to act as the model input. The model's objective is to reconstruct the original color image, with the latter serving as the ground-truth label. We use the official ImageNet-1K validation set for testing and employ the MSE to quantify the discrepancy between the predicted colors and the ground truth.

\subsection{Implementation Details}

To study the effect of prompt diversity, we vary the number of candidate prompts $K \in {\{1,16\}}$. Optimization is performed with SGD (initial learning rate 0.03) under a cosine decay schedule. The Fusion Module is first trained for up to 150 epochs on segmentation and detection, and 10 epochs on colorization. Each prompt arrangement’s MLP, with a 512-dimensional hidden layer, is then trained for 10 epochs, followed by a 10-epoch joint fine-tuning stage with selected pairs $N=2$. All experiments run on a single V100 GPU with batch size 16. The loss weight $\lambda$ is fixed to 0.6.


\section{Experiments Details}
\label{sec:ex}

\subsection{Further Explanation for Table 4 of Main Paper}
This section provides a more detailed breakdown of the experimental settings presented in Table 4 to clarify the advantages of our proposed MLP-based approach.

First, the data from rows (1) and (2) demonstrates that introducing the MLP has a negligible impact on system efficiency. The model incorporating the MLP shows minimal changes in training time, peak memory usage, inference time, and frames per second (FPS) compared to the model without it. This confirms that our design enhancement is achieved at a very low computational cost.

Furthermore, The performance evaluation under different settings is detailed as follows:
\begin{itemize}
    \item\textbf{Base (3):} This setting evaluates the baseline performance of the MAE-VQGAN model. The model was trained exclusively on the `a1' arrangement and then evaluated on all eight arrangements (a1--a8) without any fine-tuning, using the top-1 most similar prompt as support. This measures the model's inherent generalization capability to unseen arrangements.

    \item\textbf{Fuse (5):} In this configuration, after the initial training of the Fusion Module, the core MAE-VQGAN model's parameters were frozen. The Fusion Module itself was then activated and directly fine-tuned for each of the eight arrangements individually. This experiment is designed to test the effectiveness of adapting the fusion mechanism to new layouts.

    \item\textbf{MLP (7):} Here, following the Fusion Module's training, both the MAE-VQGAN and the Fusion Module were frozen. A new, separate MLP was introduced for each of the eight arrangements. Only these new MLPs were trained to handle arrangement-specific knowledge. This design effectively isolates the learning of spatial layout information into a dedicated component.

    \item\textbf{Fine-Tuning (-FT Suffix):} The "-FT" variants (Base-FT, Fuse-FT, MLP-FT) signify a comprehensive joint fine-tuning process based on a query-support swapping strategy. In these settings, all parameters of the corresponding model (Base, Fuse, or MLP) were unfrozen and fine-tuned on the target arrangements.
\end{itemize}

A key finding emerges from the comparison between the \textbf{Fuse/Fuse-FT (5)(6)} and \textbf{MLP/MLP-FT (7)(8)} results: attaching a dedicated MLP to learn arrangement-specific knowledge is significantly more effective than directly fine-tuning the Fusion Module. The latter approach can disrupt the module's primary function of semantic fusion, leading to suboptimal performance. In contrast, the MLP-based method cleanly separates the concerns of spatial encoding from semantic processing. This separation allows for more targeted and efficient learning, which ultimately results in superior overall performance.

\begin{table}
\centering
\fontsize{9}{11}\selectfont
\begin{tabular}{c ccc cc}
\toprule
\textbf{ID} & \textbf{Fusion} & \textbf{MLP} & \textbf{MAE-VQGAN} & \textbf{Seg.} & \textbf{Det.} \\
\midrule
(1) & \ding{55} & \ding{55} & \ding{55} & 46.48 & 41.13 \\
(2) & \ding{51} & \ding{55} & \ding{55} & 47.54 & 42.27 \\
(3) & \ding{55} & \ding{51} & \ding{55} & 47.56 & 42.30 \\
(4) & \ding{55} & \ding{55} & \ding{51} & 49.24 & 43.94 \\
(5) & \ding{51} & \ding{55} & \ding{51} & 50.09 & 44.83 \\
\rowcolor{lightpurple}
(6) & \ding{51} & \ding{51} & \ding{51} & \textbf{50.32} & \textbf{45.07} \\
\bottomrule
\end{tabular}
\caption{Ablation study on segmentation and detection performance with mIoU $\uparrow$.}
\label{tab:ablation_tasks_miou}
\end{table}

\subsection{Ablation Study on Joint Fine-Tuning Strategy}

To validate the effectiveness of our joint fine-tuning strategy, we conduct a comprehensive ablation study, as shown in Table~\ref{tab:ablation_tasks_miou}, which isolates the contribution of the Fusion module, the arrangement-specific MLP, and the MAE-VQGAN backbone. For segmentation, we report the mean mIoU over 4-fold cross-validation to ensure robustness. Starting from the no-tuning baseline (1), fine-tuning each component individually (2–4) yields consistent performance gains, with the MAE-VQGAN backbone showing the largest improvement, underscoring its importance in adapting inpainting capability to downstream tasks.

Building upon this, jointly tuning the Fusion module with MAE-VQGAN (5) further boosts performance, demonstrating positive synergy. The best results are obtained with the full configuration (6), where all components are unfrozen and optimized together via our query–support swapping strategy, achieving 50.32 mIoU for segmentation and 45.07 mIoU for detection. These findings confirm that full-model joint optimization, rather than partial fine-tuning, is the most effective paradigm for maximizing performance.


\section{Qualitative Results}
\label{sec:qu}

\begin{figure*}
    \centering

    \textbf{Qualitative Results for Segmentation}\par\medskip
    \begin{subfigure}{0.19\textwidth}\includegraphics[width=\linewidth]{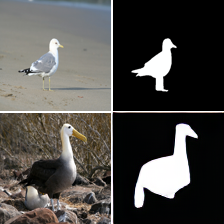}\end{subfigure}\hfill
    \begin{subfigure}{0.19\textwidth}\includegraphics[width=\linewidth]{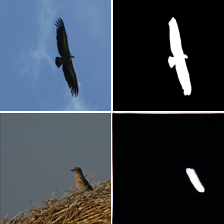}\end{subfigure}\hfill
    \begin{subfigure}{0.19\textwidth}\includegraphics[width=\linewidth]{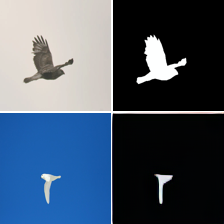}\end{subfigure}\hfill
    \begin{subfigure}{0.19\textwidth}\includegraphics[width=\linewidth]{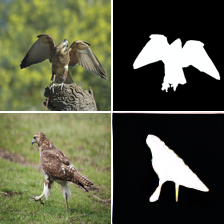}\end{subfigure}\hfill
    \begin{subfigure}{0.19\textwidth}\includegraphics[width=\linewidth]{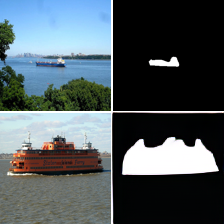}\end{subfigure}
    \vspace{2pt} \\
    \begin{subfigure}{0.19\textwidth}\includegraphics[width=\linewidth]{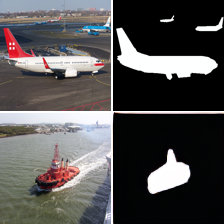}\end{subfigure}\hfill
    \begin{subfigure}{0.19\textwidth}\includegraphics[width=\linewidth]{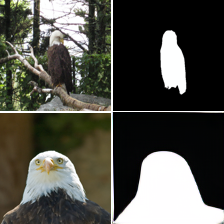}\end{subfigure}\hfill
    \begin{subfigure}{0.19\textwidth}\includegraphics[width=\linewidth]{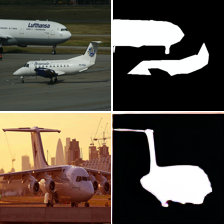}\end{subfigure}\hfill
    \begin{subfigure}{0.19\textwidth}\includegraphics[width=\linewidth]{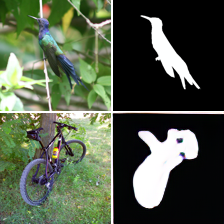}\end{subfigure}\hfill
    \begin{subfigure}{0.19\textwidth}\includegraphics[width=\linewidth]{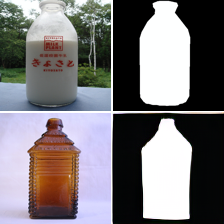}\end{subfigure}
    

    \textbf{Qualitative Results for Detection}\par\medskip
    \begin{subfigure}{0.19\textwidth}\includegraphics[width=\linewidth]{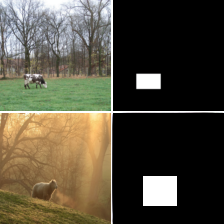}\end{subfigure}\hfill
    \begin{subfigure}{0.19\textwidth}\includegraphics[width=\linewidth]{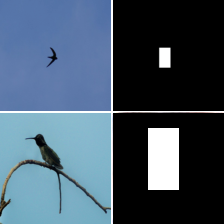}\end{subfigure}\hfill
    \begin{subfigure}{0.19\textwidth}\includegraphics[width=\linewidth]{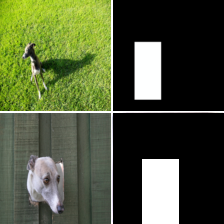}\end{subfigure}\hfill
    \begin{subfigure}{0.19\textwidth}\includegraphics[width=\linewidth]{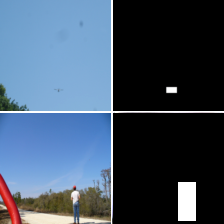}\end{subfigure}\hfill
    \begin{subfigure}{0.19\textwidth}\includegraphics[width=\linewidth]{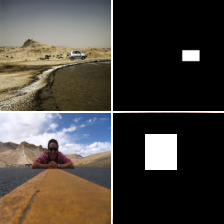}\end{subfigure}
    \vspace{2pt} \\
    \begin{subfigure}{0.19\textwidth}\includegraphics[width=\linewidth]{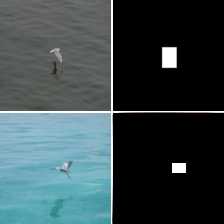}\end{subfigure}\hfill
    \begin{subfigure}{0.19\textwidth}\includegraphics[width=\linewidth]{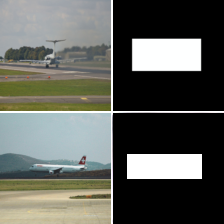}\end{subfigure}\hfill
    \begin{subfigure}{0.19\textwidth}\includegraphics[width=\linewidth]{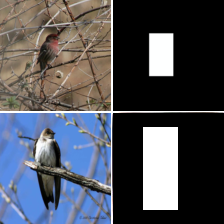}\end{subfigure}\hfill
    \begin{subfigure}{0.19\textwidth}\includegraphics[width=\linewidth]{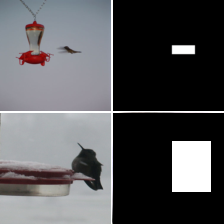}\end{subfigure}\hfill
    \begin{subfigure}{0.19\textwidth}\includegraphics[width=\linewidth]{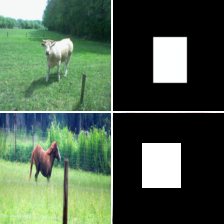}\end{subfigure}

    \textbf{Qualitative Results for Coloring}\par\medskip
    \begin{subfigure}{0.19\textwidth}\includegraphics[width=\linewidth]{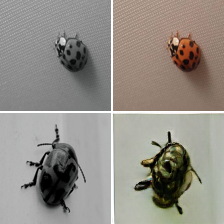}\end{subfigure}\hfill
    \begin{subfigure}{0.19\textwidth}\includegraphics[width=\linewidth]{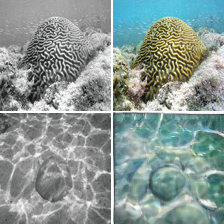}\end{subfigure}\hfill
    \begin{subfigure}{0.19\textwidth}\includegraphics[width=\linewidth]{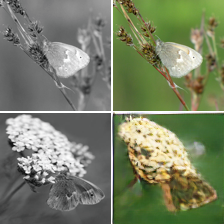}\end{subfigure}\hfill
    \begin{subfigure}{0.19\textwidth}\includegraphics[width=\linewidth]{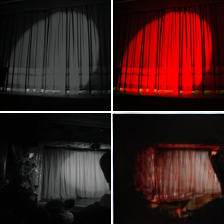}\end{subfigure}\hfill
    \begin{subfigure}{0.19\textwidth}\includegraphics[width=\linewidth]{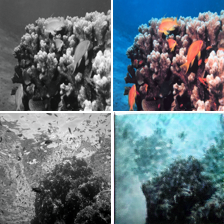}\end{subfigure}
    \vspace{2pt} \\
    \begin{subfigure}{0.19\textwidth}\includegraphics[width=\linewidth]{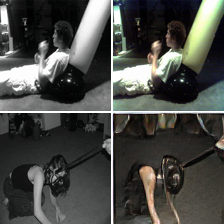}\end{subfigure}\hfill
    \begin{subfigure}{0.19\textwidth}\includegraphics[width=\linewidth]{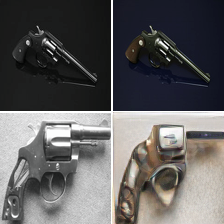}\end{subfigure}\hfill
    \begin{subfigure}{0.19\textwidth}\includegraphics[width=\linewidth]{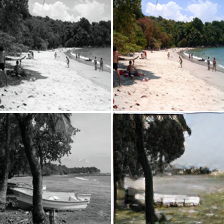}\end{subfigure}\hfill
    \begin{subfigure}{0.19\textwidth}\includegraphics[width=\linewidth]{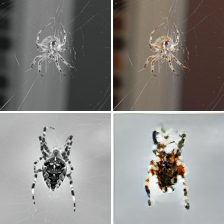}\end{subfigure}\hfill
    \begin{subfigure}{0.19\textwidth}\includegraphics[width=\linewidth]{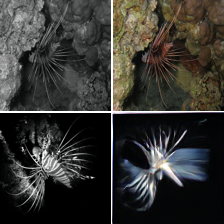}\end{subfigure}

    \caption{Qualitative results for our model on three different downstream tasks, uniformly arranged in the a1 layout. From top to bottom: Segmentation, Detection, and Coloring, showcasing the model's ability to generate high-quality, coherent outputs across diverse tasks.}
    \label{fig:qualitative_all_tasks}
\end{figure*}

In this section, we provide a selection of qualitative results to visually demonstrate the performance and versatility of our proposed model across multiple downstream tasks. The results are presented in Figure~\ref{fig:qualitative_all_tasks}.

\section{Pseudo Code}
\label{sec:code}
To provide a clear and reproducible overview of our methodology, Algorithm~\ref{alg:fuse_full} shows three stages of our method: first train an adaptive Fusion Module for creating precise contextual prompts, then decouple layout priors using lightweight, arrangement-specific MLPs, and finally employ a bidirectional fine-tuning mechanism to enhance the synergy between the fusion and inpainting models.

\begin{algorithm}
\caption{Three-stage Training of the Fusion Module $f$}
\label{alg:fuse_full}
\SetAlgoLined 
\DontPrintSemicolon 

\KwIn{Query image $I^q$, support pool $P=\{(I_i,L_i)\}_{i=1}^{N_p}$, frozen MAE-VQGAN $\varphi$, hyperparams $K$, $N$, $\lambda$}
\KwOut{Trained Fusion Module $f$}

\BlankLine
\textbf{Stage I: Training of Fusion Module $f$}\;
\For{each training iteration}{
    Extract features $I^q$, $I_i$\;
    Compute similarities $s_i=\cos(I^q,I_i)$\;
    Select Top-$K$: $\{c_k\}_{k=1}^K=\operatorname{arg\,topK}(s_i)$\;
    
    \tcp{The Fusion Module f performs attention-based feature fusion}
    Obtain fused feature and target from the module: $(I_{c_m}, L_{c_m}) \leftarrow f(I^q, \{I_{c_k}, L_{c_k}\}_{k=1}^K)$\;
    
    Construct fused pair $P_{c_m}=(I_{c_m},L_{c_m})$ and grid 
    $X_{c_m}=\begin{bmatrix}I_{c_m}&L_{c_m}\\I^q&L[\text{MASK}]\end{bmatrix}$\;
    
    Predict $T_{c_m}=\varphi(X_{c_m})$, obtain $T^L_{c_m}$\;
    Compute losses 
    $\mathcal{L}_{\text{align}}=\|I^q-I_{c_m}\|_2^2$,
    $\mathcal{L}_{\text{CE}}=-\sum_j I^L_j\log T^L_{c_m,j}$,
    $\mathcal{L}_{\text{fuse}}=\lambda\mathcal{L}_{\text{align}}+(1-\lambda)\mathcal{L}_{\text{CE}}$\;
    Update $f$ by minimizing $\mathcal{L}_{\text{fuse}}$\;
}

\BlankLine
\textbf{Stage II: Arrangement Search with Frozen $f$ and $\varphi$}\;
Freeze $f$ and $\varphi$\;
\For{each arrangement $a_i,~i=1,\dots,8$}{
    Form contextualized sample $X_{c_m}^i=\operatorname{Arrange}(P_{c_m}, I^q; a_i)$\;
    Apply adapter $\text{MLP}_i$: $Z_{c_m}^i = X_{c_m}^i + \mathrm{FFN}(X_{c_m}^i)$\;
    Decode via $\varphi$: $T^i_{c_m}=\varphi(Z_{c_m}^i)$, obtain $T^{L,i}_{c_m}$\;
    Compute loss $\mathcal{L}^i_{\text{arr}}=\text{CE}(T^{L,i}_{c_m}, I^L)$\;
    Update $\text{MLP}_i$\;
}
Select preferred arrangements $a^*$\;

\BlankLine
\textbf{Stage III: Joint Fine-tuning with Online Augmentation}\;
Unfreeze $\varphi$, $f$ and train jointly with $\text{MLP}_{a_{(*)}}$\;
Construct sample $X_{c_m}^{a_{(*)}}=\operatorname{Arrange}(P_{c_m}, I^q; a^*)$\;

\For{each training iteration}{
    Obtain output $T_{c_m}^{a_{(*)}}=\varphi(\text{MLP}_{a_{(*)}}(X_{c_m}^{a_{(*)}}))$\;
    Perform one sub-iteration without replacement\;
    Form new pair $P_{\text{new}}=(I^q, T_{c_m}^{L})$\;
    Select Top-$N$ most similar supports $Q^S=\{P_{c_n}\}_{n=1}^N \subseteq S^q$\;
    \For{$n=1$ to $N$}{
        Replace one support: 
        $S^q_{\text{new}}=(S^q \setminus \{P_{c_n}\}) \cup \{P_{\text{new}}\}$\;
        Use $S^q_{\text{new}}$ as support and $P_{c_n}$ as query to update $f$ and $\text{MLP}_{a_{(*)}}$\;
    }
    Compute $\mathcal{L}_{\text{finetune}}=\text{CE}(T_{c_m}^{a_{(*)}}, I^L)$\;
    Update all the modules\;
}
\end{algorithm}

\end{document}